\pdfoutput=1

\documentclass[11pt]{article}

\usepackage[final]{acl}
\usepackage[ruled]{algorithm2e}

\usepackage{float} 
\usepackage{multirow} 
\usepackage{array}
\usepackage{graphicx} 
\usepackage{amsmath}
\usepackage{caption} 
\usepackage{lipsum} 
\usepackage{times}
\usepackage{latexsym}

\usepackage[T1]{fontenc}

\usepackage[utf8]{inputenc}

\usepackage{microtype}

\usepackage{inconsolata}

\usepackage{graphicx}

\title{Beyond Similarity: A Gradient-based Graph Method for Instruction Tuning Data Selection}

\author{
  Yang Zhao$^{1}$, Li Du$^{2}$, Xiao Ding$^{1}\footnotemark[2]$, Yangou Ouyang$^{1}$, Hepeng Wang$^{1}$, Kai Xiong$^{1}$,\\
 Jinglong Gao$^{1}$, Zhouhao Sun$^{1}$,Dongliang Xu$^{3}$, Yang Qing$^{3}$, Dongchen Li$^{3}$, Bing Qin$^{1}$, Ting Liu$^{2}$ \\
  $^{1}$Research Center for Social Computing and Information Retrieval, Harbin Institute of Technology, China \\
  $^{2}$Beijing Academy of Artificial Intelligence, Beijing, China \\
  $^{3}$Du Xiaoman Technology (Beijing) Co., Ltd. \\
  \tt{\{yangzhao, xding, yooy, hpwang, kxiong, jlgao, zhsun, qinb,tliu\}@ir.hit.edu.cn} \\
  \tt{duli@baai.ac.cn} \\
  \tt{\{xudongliang, yangqing, lidongchen\}@duxiaoman.com}
}

\begin{document}
\maketitle
\renewcommand*{\thefootnote}
{\fnsymbol{footnote}}
\footnotetext[2]{Corresponding Author.}
\maketitle
\begin{abstract}
Large language models (LLMs) have shown great potential across various industries due to their remarkable ability to generalize through instruction tuning.
However, the limited availability of domain-specific data significantly hampers their performance on specialized tasks.
While existing methods primarily focus on selecting training data from general datasets that are similar to the target domain, they often fail to consider the joint distribution of instructions, resulting in inefficient learning and suboptimal knowledge transfer.
To address these challenges, we introduce \textbf{G2IS} (\textbf{G}radient-based \textbf{G}raph \textbf{I}nstruction \textbf{S}election), a novel method that constructs a mixed gradient-based instruction graph to capture the joint distribution and interdependencies between instructions. By accounting for the relationships between instructions, G2IS improves domain adaptation efficiency. Additionally, we propose a gradient walk algorithm to refine the data selection process, enhancing both training effectiveness and efficiency.
Our experiments demonstrate that G2IS outperforms traditional methods across various domain adaptation tasks, yielding significant performance gains, particularly in complex, data-scarce scenarios. These results underscore the potential of G2IS in advancing the development of large, domain-specific models.
\end{abstract}
\section{Introduction}
The increasing demand for personalized and domain-specific applications has driven the rapid advancement of domain-specific LLMs ~\cite{ wu2023bloomberggpt, zhang2023xuanyuan}.
Unlike general-purpose models, these models should not only develop general expertise but also continuously adapt to evolving domain knowledge.
However, the effectiveness of these models critically depends on their ability to efficiently acquire and apply relevant, domain-specific knowledge.

\begin{figure}
    \centering
    \includegraphics[width=1 \linewidth]{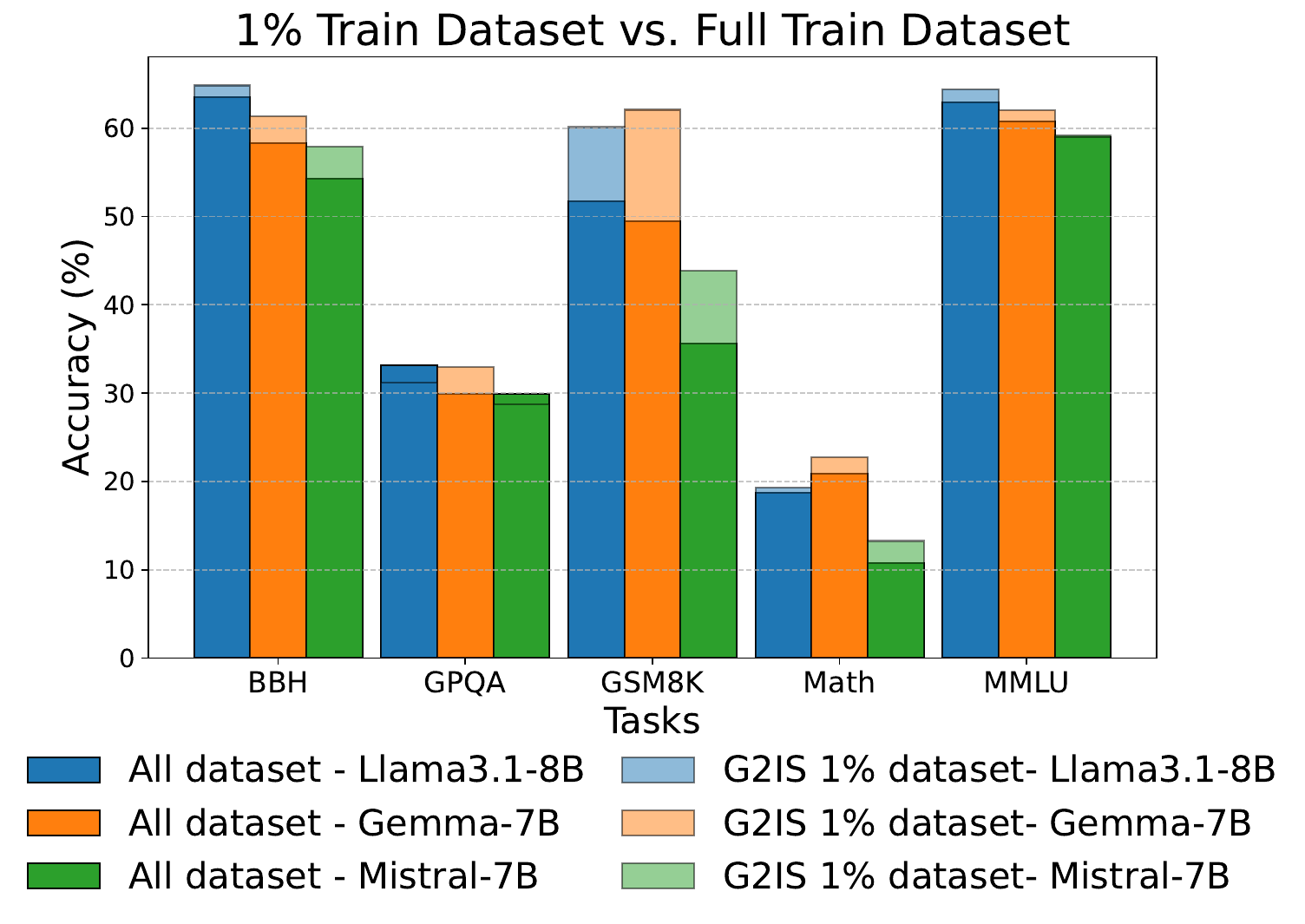}
    \caption{On the FLAN-V2 dataset, our method selects 1\% of the data and compares it with the full dataset across three models. In most tasks, our approach using only 1\% outperforms instruction tuning on the full dataset.}
    \label{fig:first}
\end{figure}

Instruction tuning has emerged as a crucial method for adapting LLMs to specialized domains through the use of curated, task-specific instruction datasets~\cite{peng2023instruction, zhang2023instruction}.
By exposing models to domain-relevant instructions, this approach significantly improves their ability to generalize across various tasks, including those requiring domain adaptation and complex reasoning.
However, a key challenge in domain-specific instruction tuning is the limited availability of high-quality annotated data. 
Additionally,concerns such as user privacy has further exacerbated the data limitations in these domain-specific areas.
To address this challenge, a promising solution is data selection, which involves identifying, from large-scale, general instruction datasets, the most relevant and impactful training samples that closely align with the target task~\cite{zhao2024supervised}. 
This process can substantially enhance the model's performance within the target domain.Therefore, the effectiveness of the data selection method is critical, as it directly influences the success of domain-specific instruction tuning.

A common approach to data selection involves choosing samples that closely resemble the validation set~\cite{xialess, joaquin2024in2core}.
However, training data typically exhibits complex interdependencies~\cite{zhaodeciphering, zhao2024beyond}, which previous methods overlook by treating relationships between data points independently.
As a result, similarity-based selection methods fail to achieve optimal performance, limiting the effectiveness of instruction tuning~\cite{hubotter2024efficiently} .
This failure to account for interdependencies hinders the construction of an optimal dataset, restricting the model’s ability to generalize.
Fundamentally, data selection aims to identify the data from the training set that possesses the capabilities required to complete the target task.
Instruction data distributions form joint distributions, where relationships between instructions should be considered~\cite{zhao2024beyond}.
However,earlier methods~\cite{xialess, joaquin2024in2core} failed to account for this joint distribution, leading to suboptimal data selection.

To address these challenges, we propose G2IS, an innovative approach leveraging gradient-based knowledge representation for more efficient data selection.
G2IS employs a mixed instruction gradient-based graph that models the complex relationships between instructions.
This graph captures the joint distribution of instruction data, enabling more effective data selection.
Model gradients capture the informational content of training samples~\cite{park2023trak, jainimproving, zhao2024supervised}, influencing parameter updates and implicitly encoding how each sample contributes knowledge to the model~\cite{hammoudeh2024training}.
Building on this, we enhance the robustness of the validation set by applying Principal Component Analysis (PCA)~\cite{kurita2021principal} to the gradients, extracting core knowledge representations that guide data selection.
For the training set, we introduce a gradient walk algorithm that refines sample selection by leveraging these gradients.
This approach considers the joint distribution of instruction data, progressively selecting samples that align with the core knowledge identified in the validation set.
By structuring the data selection process this way, we ensure that the selected training data is efficient and aligned with key knowledge, improving the overall quality of the training process.

We validate G2IS across multiple domain benchmarks.
As shown in Figure \ref{fig:first}, our method achieves exceptional performance with only 1\% of the training data, outperforming full-data instruction tuning on most tasks.
Notably, on the GSM8K dataset using the Gemma-7B model, our approach improves \textbf{accuracy by 12.66\%} compared to full instruction tuning.
These results highlight the efficiency of G2IS in reducing data requirements while maintaining or even surpassing baseline performance.

\section{Gradient-based Knowledge Representation}
During instruction tuning, LLMs update their parameters through gradient-based optimization, making gradients a natural representation of model knowledge.
Gradients reflect the influence of individual training samples on parameter updates, revealing which data points contribute most to model learning.
In this section, we explain how gradients are computed for the training and validation sets and how they are used to construct a \textbf{gradient-based graph} for efficient data selection. 
In the next section, we introduce how to utilize the Gradient-based Graph for Instruction Selection.

\subsection{Gradient-based Knowledge Representation}
Gradients not only drive parameter updates during instruction tuning but also implicitly encode knowledge about how training data influences model learning~\cite{choe2024your}.
By capturing the directional influence of each sample on parameter updates, gradients naturally reveal the relationships between instructions.
Unlike static embeddings or similarity-based methods, gradients provide a dynamic, task-sensitive representation of knowledge~\cite{hammoudeh2024training}, effectively capturing both similarities and deeper interdependencies within the training data.

This can be understood through a first-order Taylor expansion of the loss function, where the model parameters update as:
\begin{equation}
\theta' = \theta - \eta \nabla \mathcal{L}(z, \theta), 
\end{equation}
where \(\eta\) represents the learning rate, and the gradient \( \nabla \mathcal{L}(z, \theta) \) determines how each sample modifies the model, encapsulating its contribution to learning.
Thus, the relationships between gradients reflect the dependencies between instruction data, with the similarity of instruction gradients enabling joint modeling and revealing the complex relationships within instruction tuning data.

\begin{figure*}
    \centering
    \small
    \includegraphics[width=0.9 \linewidth]{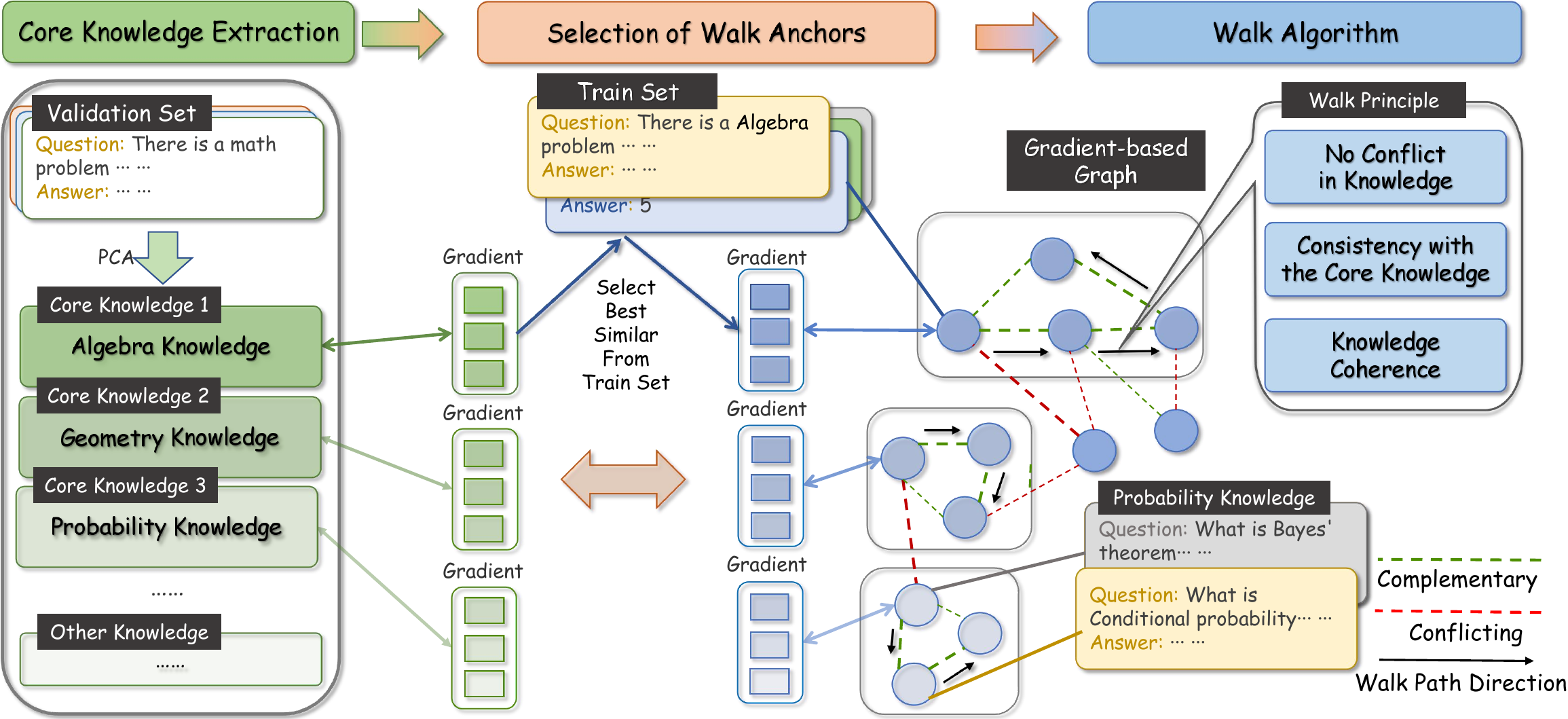}
    \caption{The left side illustrates the core knowledge extracted from the validation set. In the center, anchor selection for the gradient walk is performed by identifying the most similar data points from the training set, based on the core knowledge. These anchors are then used to conduct a gradient walk in the graph (right side), ensuring three key conditions: No Conflict in Knowledge, Consistency with Core Knowledge, and Knowledge Coherence. Finally, a gradient subgraph is selected in the lower-right corner, representing a subset of knowledge from the training set (e.g., probabilistic knowledge).}
    \label{fig:main}
\end{figure*}
\subsection{Gradient Computation on Training and Validation Sets}

To accurately capture the knowledge representation of different data samples, we compute gradients separately for both the training and validation sets.

For the training set, we define the \textbf{momentum-adjusted gradient} \( \nabla \Gamma(z, \theta_t) \), which represents the effective gradient used for parameter updates in modern LLMs trained with the Adam optimizer~\cite{kingma2014adam}.
Unlike standard gradients, it incorporates both first-order and second-order momentum terms, providing a more accurate representation of how training samples contribute to model optimization. 
Directly computing raw gradients without considering momentum effects can lead to imprecise knowledge representations.
To address this, we use the \textbf{warmup method} proposed by~\citet{xialess, liu2024less}, where the model is pre-trained on a small subset of instructions to initialize the optimizer's momentum states.
This ensures that \( \nabla \Gamma(z, \theta_t) \) reflects the actual optimization dynamics, improving data selection precision and instruction representation.

For the validation set, we compute the first-order gradient of the loss function with respect to model parameters for each sample.
This gradient directly measures how each sample influences parameter updates. To avoid momentum interference, prior research~\cite{xialess} showed that using stochastic gradient descent (SGD) for validation gradients and Adam for training gradients improves data selection accuracy.
Therefore, we compute these gradients using standard SGD without modifications.

Given the high computational cost of calculating full gradients for all model parameters, we leverage LoRA~\cite{hu2021lora} to compute gradients within LoRA layers, significantly reducing memory overhead while preserving key gradient information.
Additionally, we apply \textbf{Random Projection}~\cite{johnson1984extensions, park2023trak} dimensionality reduction techniques to efficiently extract low-dimensional gradient features, ensuring computational efficiency without sacrificing essential knowledge representation.

\subsection{Gradient-based Graph Construction}

After obtaining gradient representations, we construct a structured \textbf{gradient-based graph}, consisting of nodes \( N_z \) and edges \( R_{ij} \).
Each node \( N_z \) in the graph is defined as:
\begin{equation}
\textstyle N_z = \nabla \Gamma(z, \theta_t),
\end{equation}
where \( \nabla \Gamma(z, \theta_t) \) encapsulates the sample’s contribution to model updates, effectively encoding its role in the instruction tuning process.
Edges \( R_{ij} \) between nodes \( z_i \) and \( z_j \) are weighted by their cosine similarity:
\begin{equation}
\textstyle R_{ij} = \cos\left( \nabla \Gamma(z_i, \theta_t), \nabla \Gamma(z_j, \theta_t) \right),
\end{equation}
where higher values indicate stronger alignment in their learning impact, and negative values suggest potential conflicts.
This structure captures not only direct similarities but also complementary and conflicting relationships between samples.
By modeling the training data as a gradient-based graph, we capture the interdependencies overlooked by traditional similarity-based methods, laying the foundation for further capturing joint training samples that meet specific task requirements.

\section{Gradient-based Graph for Instruction Selection}

Building on the gradient-based graph, as shown in Figure \ref{fig:main}, we apply PCA to reduce noise in the validation set and extract core knowledge.
Using this core knowledge, we identify the walk anchor to select the most relevant training data for the validation set, achieved through a walk algorithm for data selection.

\subsection{Core Knowledge Extraction from the Validation Set and Selection of Walk Anchors}
\label{Core Knowledge Extraction from the Validation Set}

The validation set, which is typically small and designed to resemble the test set, is often assumed to be independent and identically distributed in previous studies~\cite{xialess, joaquin2024in2core}, where the average gradient is used as a proxy for the core knowledge it encapsulates.
However, the main goal of the validation set is not only to represent the overall knowledge but also to capture the essential capabilities required to solve specific tasks.
To more accurately extract this core knowledge, we employ PCA on the gradient distribution of the validation set.
PCA identifies the principal components of the gradients, which correspond to the most critical task-related capabilities.
Based on these principal components, we select the most relevant training samples as anchors for the walk algorithm, using them as the starting point for the gradient walk.
Unless otherwise stated, we use the first 50\% of principal components in our experiments, as they capture the majority of the variance in the data, ensuring that the selected samples align with the core knowledge required for the task.

\begin{table*}[t!]
\centering
\small
\renewcommand{\arraystretch}{1.6} 
\begin{tabular}{c@{\hspace{1.5pt}} c@{\hspace{1.5pt}} c@{\hspace{2.5pt}} c@{\hspace{2.5pt}} c@{\hspace{2.5pt}} c@{\hspace{2.5pt}} c@{\hspace{2.5pt}}  c@{\hspace{2.5pt}} c@{\hspace{2.5pt}} c@{\hspace{2.5pt}} c@{\hspace{2.5pt}} c@{\hspace{2.5pt}} c@{\hspace{2.5pt}} c@{\hspace{2.5pt}} c@{\hspace{2.5pt}} c@{\hspace{2.5pt}} c@{\hspace{2.5pt}}}
\hline
&&\multicolumn{5}{c}{Llama3.1-8B} &\multicolumn{5}{c}{Gemma-7B} &\multicolumn{5}{c}{Mistral-7B} 
\\\hline
ratio&Ins&BBH&GPQA&GSM8K&Math&MMLU&BBH&GPQA&GSM8K&Math&MMLU&BBH&GPQA&GSM8K&Math&MMLU\\
\hline
 & BERT & 63.05 & 30.55 & 56.79 & 20.22 & 61.87  & 49.79 & 34.22 & 53.15 & 20.92 & 60.75 & 56.77 & 24.85 & 43.44 & 13.30 & 58.30\\
1\% & LESS & 63.46 & 29.94 & 60.65 & 18.66 & 63.15  & 58.95 & 30.96 & 58.45 & 20.02 & 59.85 & 57.93 & 27.49 & 44.05 & 14.04 & 55.26\\
 & G2IS & \textbf{64.78} & \textbf{31.57} & \textbf{62.02} & \textbf{20.96} & \textbf{63.42}  & \textbf{60.25} & \textbf{35.03} & \textbf{61.64} & \textbf{22.88} & \textbf{62.17} & \textbf{58.59} & \textbf{29.44} & \textbf{46.40} & \textbf{14.68} & \textbf{58.64}\\
\hline
 & BERT & 59.65 & 26.68 & 50.95 & 20.46 & 62.33  & 44.39 & 31.57 & 55.80 & 22.16 & 59.85 & 53.72 & 26.27 & 45.79 & 14.02 & 58.00\\
5\% & LESS & 64.17 & 29.94 & 63.08 & 18.90 & 62.87  & 58.65 & 31.16 & 60.58 & 20.26 & 59.86 & 56.06 & 28.51 & 52.24 & 13.78 & 58.20\\
 & G2IS & \textbf{65.07} & \textbf{34.22} & \textbf{64.06} & \textbf{21.00} & \textbf{63.36}  &  \textbf{59.38}& \textbf{34.42} & \textbf{62.70} & \textbf{23.00} & \textbf{62.13} & \textbf{57.40} & \textbf{31.57} & \textbf{54.44}& \textbf{14.14} & \textbf{59.28}\\
\hline
All & & 64.71 & 29.74 & 58.30 & 20.26 & 62.75   & 58.12 & 28.72 & 63.31 & 23.64 & 59.66 & 56.95 & 29.74 & 52.99 & 18.52 & 58.04\\
\hline  
\hline
ratio&COT&BBH&GPQA&GSM8K&Math&MMLU&BBH&GPQA&GSM8K&Math&MMLU&BBH&GPQA&GSM8K&Math&MMLU\\
\hline
 & BERT & 63.68 & 28.51 & 53.90 & 16.36 & 63.04  & 58.50 & 28.92 & 57.77 & 21.08 & 59.61 & 56.11 & 23.42 & 40.79 & 12.30 & 57.87\\
1\% & LESS & 64.29 & 27.90 & 57.70 & 18.54 & 63.42  & 59.47 & 32.18 & 57.47 & 21.28 & 59.43 & 56.38 & 27.49 & 44.81 & 11.54 & 57.04\\
 & G2IS & \textbf{65.66} & \textbf{32.59} & \textbf{62.70} & \textbf{21.38} & \textbf{64.22}  & \textbf{61.60} & \textbf{33.20} & \textbf{63.76} & \textbf{23.44} & \textbf{62.34} & \textbf{58.04} & \textbf{28.72} & \textbf{49.20} & \textbf{12.64} & \textbf{59.09}\\
\hline
 & BERT & 63.60 & 29.33 & 61.11 & 17.60 & 62.48  & 56.77 & 29.94 & 60.12 & 21.40 & 58.94 & 54.42 & 28.51 & 44.58 & 11.68 & 56.22\\
5\% & LESS & 62.76 & 29.74 & 60.73 & 17.74 & 63.02  & 59.15 & 29.74 & 60.88 & 21.08 & 59.31 & 56.87 & 27.29 & 48.45 & 11.32 & 58.05\\
 & G2IS  & \textbf{65.14} & \textbf{30.55} & \textbf{62.47} & \textbf{20.68} & \textbf{64.01} & \textbf{60.44} & \textbf{35.23} & \textbf{64.90} & \textbf{23.10} & \textbf{61.85} & \textbf{57.68} & \textbf{28.92} & \textbf{49.41} & \textbf{13.36} & \textbf{59.48}\\
\hline
All & & 60.18 & 30.35 & 60.58 & 17.06 & 60.35   & 56.58 & 29.53 & 60.35 & 19.46 & 59.01 & 54.58 & 27.29 & 51.78 & 10.42 & 57.79\\
\hline
\hline
ratio&FLAN&BBH&GPQA&GSM8K&Math&MMLU&BBH&GPQA&GSM8K&Math&MMLU&BBH&GPQA&GSM8K&Math&MMLU\\
\hline
 & BERT & 64.03 & 29.53 & 50.99 & 15.58 & 63.50  & 59.88 & 30.75 & 52.16 & 19.14 & 59.95 & 55.60 & 23.01 & 39.50 & 12.50 & 56.89\\
1\% & LESS & 64.81 & 28.72 & 51.55 & 14.92 & 60.79  & 57.76 & \textbf{32.99} & 56.18 & 21.68 & 59.92 & 56.47 & 26.07 & 37.60 & 12.38 & 51.64\\
   & G2IS & \textbf{64.84} & \textbf{31.16} & \textbf{60.12} & \textbf{19.30} & \textbf{64.36 } & \textbf{61.39} & \textbf{32.99} & \textbf{62.09} & \textbf{22.70} & \textbf{62.06} & \textbf{57.89} & \textbf{28.72} & \textbf{43.90} & \textbf{13.26} & \textbf{59.19}\\
\hline
   & BERT & 62.94 & 29.53 & 50.57 & 17.26 & 62.90  & 59.15 & 30.96 & 52.54 & 17.62 & 60.31 & 54.83 & 26.68 & 37.07 & 12.98 & 53.87\\
5\% & LESS & 62.76 & 29.33 & 52.31 & 18.32 & 63.77  & 59.21 & 30.75 & 54.44 & 21.46 & 60.72 & 55.55 & 26.68 & 38.36 & 11.04 & 58.74\\
  & G2IS  & \textbf{65.27} & \textbf{30.96} & \textbf{62.40} & \textbf{20.40} & \textbf{64.03 } & \textbf{60.70} & \textbf{34.01} & \textbf{63.68} & \textbf{22.44} & \textbf{61.78 } & \textbf{59.13} & \textbf{27.29} & \textbf{48.98} & \textbf{13.46} & \textbf{59.77}\\
\hline
All & & 63.52 & 33.2 & 51.71 & 18.70 & 62.88  &  58.29 & 29.94 & 49.43 & 20.88 & 60.80 & 54.29 & 29.94 & 35.63 & 10.74 & 58.99\\
\hline
\end{tabular} 
\caption{  
Performance comparison of data selected from the top 1\% and 5\% most beneficial samples for enhancing BBH, GPQA, GSM8K, Math, and MMLU tasks.Experiments were conducted using Llama3.1-8B, Gemma-7B, and Mistral-7B v0.3, fine-tuned on Infinity-Instruct, COT, and FLAN-v2 for single-task objectives.}
\label{table:main}
\end{table*}




\subsection{Data Selection with a Walk Algorithm on the Gradient-based Graph}
\label{Data Selection Using a Walk Algorithm on the gradient-based graph}

To ensure that the selected training data effectively supports the target task, we adopt a structured data selection process based on a gradient-based graph, rather than relying solely on similarity-based methods.
By utilizing the gradient graph constructed in the previous section, we capture both complementary and conflicting knowledge structures.

The gradient walk algorithm begins with a anchor that closely resembles the validation set.
Using the weights corresponding to the principal components of the validation set, the number of training instructions required for each core knowledge component is determined.
The algorithm then expands the instruction set by selecting data that contributes positively to model learning while maintaining consistency with the validation set.

The data selection process is governed by three heuristic principles. First, \textbf{No Conflict in Knowledge} ensures that new samples do not introduce contradictions. The similarity between the new node \( z \) and all existing nodes \( s \) should be non-negative. Second, \textbf{Consistency with Core Knowledge} prevents the selected training data from deviating significantly from the core knowledge by ensuring that the cosine similarity between the instruction set and Core Knowledge \( K_\mathcal{V} \) remains below a threshold \( \delta \). Third, \textbf{Knowledge Coherence} ensures that, building on the previous two principles, the selected data is most similar to the most recent instruction \( s^* \).
Formally, the data selection process is defined as:
\begin{equation}
z^* = \underset{z \in \mathcal{Z}}{\arg\max} \, \cos\left( \nabla \Gamma(z, \theta_t), \nabla \Gamma(s^*, \theta_t) \right),
\end{equation}
subject to the following constraints:
\begin{equation}
\cos\left( \Gamma(z, \theta_t), \Gamma(s, \theta_t) \right) \geq 0, \quad \forall s \in \mathcal{S},
\end{equation}
\begin{equation}
\begin{aligned}
\big| \cos\left( \Gamma(\mathcal{S} \cup \{z\}, \theta_t), K_{\mathcal{V}} \right) \big| &\geq\\
 \delta \, \big| \cos\left( \Gamma(\mathcal{S}, \theta_t), K_{\mathcal{V}} \right) \big|,
\end{aligned}
\end{equation}
where \( z^* \) is the next selected node, \( \mathcal{Z} \) is the nodes in the gradient-based graph, \( \mathcal{S} \) is the current instruction set, and $K_{\mathcal{V}}$ is a core knowledge from the validation set.
The function \( \text{cos}(\Gamma(\mathcal{S}, \theta_t), K_{\mathcal{V}}) \) represents the cosine similarity between the instruction set gradient and the core knowledge from the validation set.
\( s^* \) refers to the most recently added node in the instruction set, ensuring new selections align with the evolving training set.
The parameter \(\delta\) acts as a threshold controlling the allowable difference in gradient similarity.
Unless specified otherwise, we set \(\delta = 0.8\) to balance diversity in selected data with alignment to the validation set.

For a detailed description of the algorithm, refer to Appendix~\ref{alg:gradient-walk}. 
By incorporating the joint distribution between instructions, it maximizes consistency with both the validation set and its inherent knowledge, while simultaneously avoiding conflicts, thereby enhancing the model's learning efficiency.






\section{Experiment}

To validate our method, we focus on two key areas: data selection for single-task optimization and gradient-based selection for multi-task instruction tuning. By comparing with baseline methods, we demonstrate the robustness and generalizability of our approach in both settings.

\subsection{Experimental Setup and Baselines}
\label{Model and Data Preparation}

To ensure a fair evaluation, we selected state-of-the-art language models: Llama3.1-8B~\cite{dubey2024llama}, Gemma-7B~\cite{team2024gemma}, and Mistral-7Bv0.3~\cite{jiang2023mistral}. Details of the training and testing data are provided in Appendix~\ref{appendix:data}.

We compared our method with two baselines: Less~\cite{xialess}, a leading data selection approach, and Sentence-BERT~\cite{reimers-2019-sentence-bert}, a widely used training data selection method. Training was performed using the Llama-Factory~\cite{zheng2024llamafactory}, and model performance was evaluated with the Harness~\cite{eval-harness}, which provides a standardized assessment methodology. Further details on experimental configurations are in Appendix~\ref{appendix:train}.
\subsection{Optimizing Data Selection for Single-Task Performance}
\label{Single-Objective Data Selection}

Table \ref{table:main} presents a comparative analysis of G2IS against baseline methods across multiple benchmark datasets.
The results show that G2IS consistently outperforms Less and Sentence-BERT, regardless of whether 1\% or 5\% of the training data is selected.
Notably, our method excels on multi-task datasets like FLAN, which is consistent with~\citet{wang2023far}, suggesting that targeted data selection outperforms full-dataset instruction tuning.
Compared to the other two datasets, FLAN-V2 covers a broader range of tasks, better simulating the process of extracting specific domain instructions from large-scale data.
Furthermore, G2IS demonstrates significant improvements on complex reasoning tasks like GSM8K and BBH, showcasing the advantages of using a gradient-based graph structure to jointly model instruction tuning data, leading to better performance on complex tasks.

A particularly noteworthy observation is that selecting just 1\% of the training data often yields better performance than using 5\% or even the entire dataset.
This suggests that a small subset of highly relevant data plays a crucial role in enhancing task-specific performance, while the presence of excessive or irrelevant data can introduce noise and hinder the model’s generalization capabilities.

Additionally, our findings show that models trained on the full dataset generally underperform compared to those trained on the top 1\% and 5\% subsets selected by G2IS. 
This further validates that blindly instruction tuning on the entire dataset does not yield optimal results.
Instead, carefully curated, high-quality subsets align better with task requirements, especially for complex or specialized tasks. 
These results are consistent with the conclusions drawn by~\citet{tsai2024code} and~\citet{zhou2024lima}. 
They confirm the effectiveness of gradient-based, graph-driven data selection in enhancing model performance and training efficiency.

\subsection{Enhancing Multi-Task Instruction tuning with Gradient-based Data Selection}
\label{Multi-Objective Data Selection}

To further assess the robustness of our method, we evaluated its performance in multi-objective optimization by combining the validation sets of GSM8K and BBH.
GSM8K focuses on mathematical reasoning, requiring step-by-step problem-solving, while BBH is a general complex reasoning dataset.
Although distinct, these tasks share significant structural similarities, making them suitable for multi-task optimization.

As shown in Table \ref{table:multi}, traditional methods like Less struggle with multi-task optimization, showing performance degradation in multi-task scenarios.
In contrast, G2IS demonstrates robustness in handling multiple objectives, consistently delivering strong results across both.
These findings confirm that our gradient-based graph approach effectively balances conflicting objectives, enabling efficient multi-task instruction tuning.

\begin{table}[t]
\centering
\small
\renewcommand{\arraystretch}{1.6} 
\begin{tabular}{c@{\hspace{1pt}} c@{\hspace{5pt}} c@{\hspace{3pt}} c@{\hspace{6pt}} c@{\hspace{6pt}} c@{\hspace{3pt}} c@{\hspace{6pt}} c@{\hspace{6pt}} c@{\hspace{3pt}} c@{\hspace{6pt}}}
\hline
\textbf{ratio}&&\multicolumn{2}{c}{\textbf{Llama3.1-8B}} 
&\multicolumn{2}{c}{\textbf{Gemma-7B}} 
&\multicolumn{2}{c}{\textbf{Mistral-7B}} \\
\hline
&FLAN&BBH&GSM8K&BBH&GSM8K&BBH&GSM8K\\
\hline
&BERT&63.74&52.16&60.82&53.15&56.37&37.53\\
1\%&LESS&64.05&53.53&56.87&53.53&56.70&39.95\\
&G2IS&\textbf{65.01}&\textbf{58.98}&\textbf{61.05}&\textbf{62.32}&\textbf{58.45}&\textbf{43.44}\\
\hline
&BERT&62.40&50.27&60.07&50.57&56.09&35.10\\
5\%&LESS&59.07&46.55&56.14&51.10&53.89&34.50\\
&G2IS&\textbf{65.81}&\textbf{57.85}&\textbf{61.36}&\textbf{59.97}&\textbf{58.09}&\textbf{46.40}\\
\hline
All&&63.52&51.71&58.29&49.43&54.29&35.63\\
\hline
&COT&BBH&GSM8K&BBH&GSM8K&BBH&GSM8K\\
\hline
&BERT&63.23&55.88&60.57&51.78&55.64&39.20\\
1\%&LESS&63.69&58.91&59.35&60.05&56.01&44.73\\
&G2IS&\textbf{64.60}&\textbf{60.12}&\textbf{61.36}&\textbf{62.93}&\textbf{58.93}&\textbf{49.51}\\
\hline
&BERT&63.40&59.21&56.54&58.38&55.55&46.70\\
5\%&LESS&63.46&57.16&58.65&56.48&55.24&45.03\\
&G2IS&\textbf{63.69}&\textbf{60.88}&\textbf{60.74}&\textbf{59.29}&\textbf{58.04}&\textbf{47.08}\\
\hline
All&&60.18&60.58&56.58&60.35&54.58&51.78\\
\hline
\end{tabular}
\caption{Performance of different methods in selecting data that simultaneously improves BBH and GSM8K complex reasoning tasks}
\label{table:multi}
\end{table}

\begin{table*}[h]
\small
\centering
\renewcommand{\arraystretch}{1.8} 
\begin{tabular}{>{\centering\arraybackslash}m{1.4cm}  >{\centering\arraybackslash}m{0.5cm} c@{\hspace{5pt}} c@{\hspace{3pt}} c@{\hspace{3pt}} c@{\hspace{3pt}} c@{\hspace{3pt}} c@{\hspace{3pt}} c@{\hspace{3pt}}| c@{\hspace{3pt}} c@{\hspace{3pt}} c@{\hspace{3pt}} c@{\hspace{3pt}} c@{\hspace{3pt}} c@{\hspace{3pt}}}
\hline
\multirow{2}{*}{\textbf{Model}}& \multirow{2}{*}{\textbf{ratio}} &\multirow{2}{*}{\textbf{w/o}} &&&\multicolumn{2}{c}{\textbf{COT}}&&&&& \multicolumn{2}{c}{\textbf{FLAN}}\\
\cline{4-15}
&&& BBH & GPQA & GSM8K & Math & MMLU & Avg  & BBH & GPQA & GSM8K & Math & MMLU & Avg \\
\hline
\multirow{6}{*}{Llama3.1-8B} & \multirow{3}{*}{1\%} &w/o graph & 65.64 & 30.55 & 57.85 & 20.1 & 61.42 & 0.95  & 64.03 & 30.55 & 58.15 & 18.96 & 61.6 & 0.97 \\
\cline{3-15}
& & w/o gradient & 64.57 & 32.53 & 58.91 & 20.24 & 63.94 & 0.97 &  64.75 & 30.14 & 57.01 & 19.12 & 63.84 & 0.98 \\
\cline{3-15}
& & G2IS & \textbf{65.66} & \textbf{32.59} & \textbf{62.7} & \textbf{21.38} & \textbf{64.22} & \textbf{1.0} &  \textbf{64.84} & \textbf{31.16} & \textbf{60.12} & \textbf{19.3} & \textbf{64.36} & \textbf{1.0} \\
\cline{2-15}
&\multirow{3}{*}{5\%} &w/o graph &64.44&30.35&55.72&18.22&58.89&0.94 &64.37&28.11&53.53&17.4&60.81&0.91 \\
\cline{3-15}
&&w/o gradient&64.20 &28.11&54.51&16.44&63.71&0.91 &63.35&28.31&55.04&17.66&64&0.93\\ 
\cline{3-15}
&&G2IS&\textbf{65.14}&\textbf{30.55}&\textbf{62.47}&\textbf{20.68}&\textbf{64.01}&\textbf{1.0}&\textbf{65.27}&\textbf{30.96}&\textbf{62.4}&\textbf{20.4}&\textbf{64.03}&\textbf{1.0}\\
\hline
\multirow{6}{*}{Gemma-7B}& \multirow{3}{*}{1\%}&w/o graph &58.61 &31.77 &59.89 &21.46 &60.92 &0.95 &60.88 &31.16 &61.41 &21.68 &59.67 &0.97\\
\cline{3-15}
&&w/o gradient &58.93 &32.79 &62.62 &22.5 &61.45 &0.98 &59.48 &30.14 &58.38 &22.58 &61.71 &0.96\\ 
\cline{3-15}
&&G2IS &\textbf{61.6} &\textbf{33.2} &\textbf{63.76} &\textbf{23.44} &\textbf{62.34} &\textbf{1.00}  &\textbf{61.39} &\textbf{32.99} &\textbf{62.09} &\textbf{22.7} &\textbf{62.06} &\textbf{1.00} \\
\cline{2-15}
&\multirow{3}{*}{5\%}&w/o graph &58.15 &31.16 &56.1 &20.2 &60.56 &0.91 &59.75 &33.6 &56.25 &21.48 &60.74 &0.96 \\
\cline{3-15}
&&w/o gradient &59.48 &32.59 &59.29 &21.56 &60.60 &0.95 &59.67 &30.96 &58.07 &21.26 &61.35 &0.95 \\
\cline{3-15}
&&G2IS &\textbf{60.44} &\textbf{35.23} &\textbf{64.9} &\textbf{23.1} &\textbf{61.85} &\textbf{1.00}  &\textbf{60.7} &\textbf{34.01} &\textbf{63.68} &\textbf{22.44} &\textbf{61.78} &\textbf{1.00} \\
\hline

\multirow{6}{*}{Mistral-7B}& \multirow{3}{*}{1\%}&w/o graph &57.06 &28.52 &43.52 &12.46 &59.02 &0.97  &57.72 &28.33 &39.8 &11.92 &58.31 &0.95 \\
\cline{3-15}
&&w/o gradient &55.83 &25.87 &43.52 &12.56 &58.71 &0.96  &57.27 &27.49 &\textbf{43.9} &12.82 &58.3 &0.98 \\
\cline{3-15}
&&G2IS &\textbf{58.04} &\textbf{28.72} &\textbf{49.2} &\textbf{12.64} &\textbf{59.09} &\textbf{1.00} &\textbf{57.89} &\textbf{28.72} &\textbf{43.9} &\textbf{13.26} &\textbf{59.19} &\textbf{1.00}\\
\cline{2-15}
&\multirow{3}{*}{5\%}&w/o graph &57.28 &28.51 &42.3 &12.44 &59.01 &0.95  &57.1 &27.24 &44.35 &12.34 &58.99 &0.95 \\
\cline{3-15}
&&w/o gradient &57.4 &28.11 &44.96 &11.78 &57.26 &0.94  &57.99 &26.99 &45.11 &12.46 &58.45 &0.96 \\
\cline{3-15}
&&G2IS &\textbf{57.68} &\textbf{28.92} &\textbf{49.41} &\textbf{13.36} &\textbf{59.48} &\textbf{1.00} &\textbf{59.13} &\textbf{27.29} &\textbf{48.98} &\textbf{13.46} &\textbf{59.77} &\textbf{1.00} \\
\hline

\end{tabular}
\caption{  
An ablation study was conducted on the COT and FLAN datasets using Llama3.1-8B, Gemma-7B, and Mistral-7B models."w/o graph" refers to a variant with no graph structure,where training samples are selected solely based on principal component analysis."w/o gradient" replaces the gradient-based representation with a Sentence-BERT-based similarity measure for comparison.The experiment evaluates data selection for BBH, GPQA, GSM8K, Math, and MMLU tasks."Avg" represents the average accuracy of each method relative to our approach.}
\label{table:ablation}
\end{table*}

\begin{figure}[h]
    \centering
    \includegraphics[width=0.9 \linewidth]{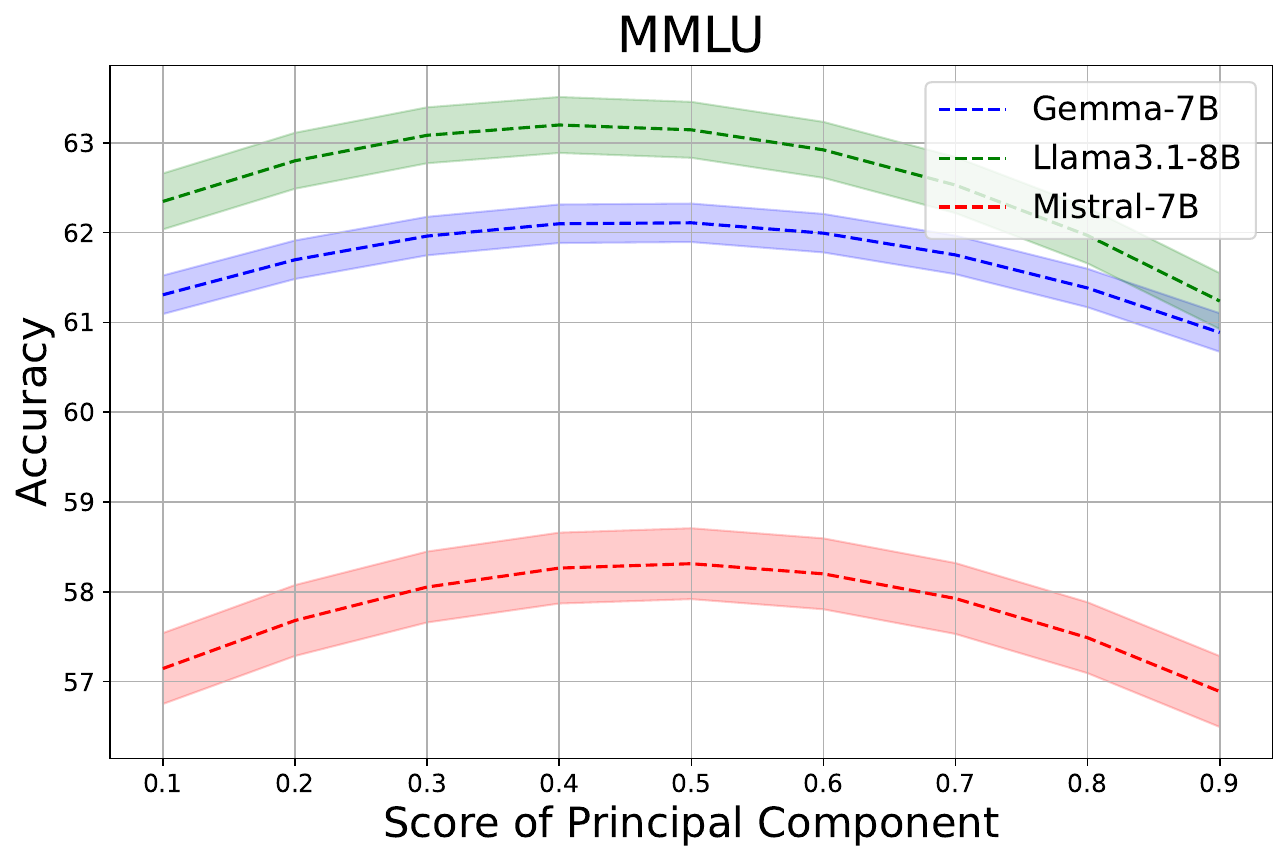}
    \small
     \includegraphics[width=0.9 \linewidth]{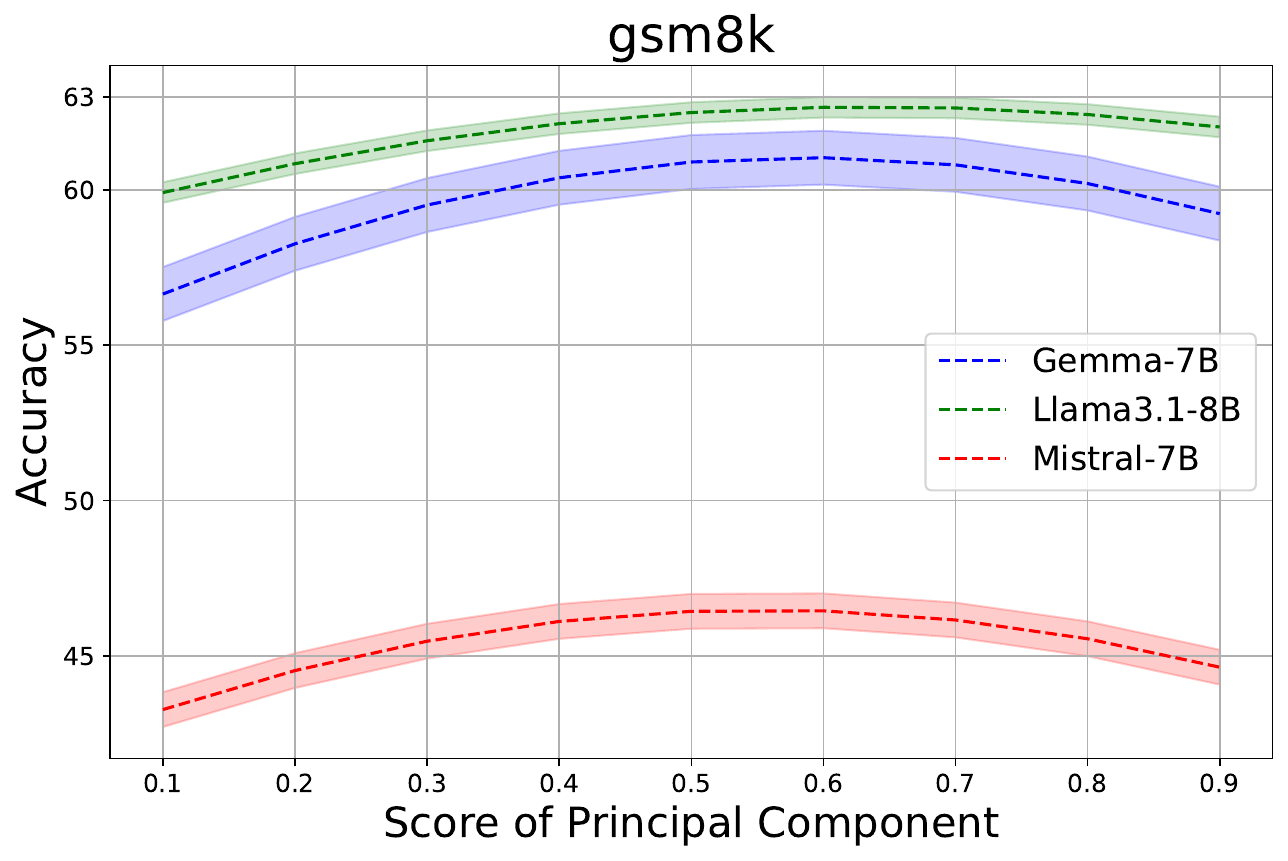}
     \small
    \caption{
    On the Infinity-Instruct dataset, we utilize the gradient walk algorithm based on principal components of knowledge extracted from different proportions of the validation set.The results of training after selecting data that enhances MMLU (upper) and GSM8K (lower) are presented.}
    \label{fig:pca}
\end{figure}

\section{Ablation Study}
Building on our experimental results, we conduct ablation studies to identify the key factors contributing to the effectiveness of our method. Specifically, we examine two aspects: (1) the impact of varying principal component ratios on task performance by selecting different proportions of principal components and analyzing their influence on the results, and (2) the role of components in the structured gradient-based graph. To explore this, we perform controlled experiments to isolate the effects of each factor. We compare our method with variants that either use PCA without the graph structure or replace gradient-based representations with BERT-based semantic similarity measures.

\subsection{Impact of Principal Component Ratios}
To investigate the impact of principal component ratios on task performance, we select the top 1\% of data from the Infinity-Instruct dataset and evaluate it using the MMLU and GSM8K benchmarks. 
As shown in Figure \ref{fig:pca}, the optimal principal component ratio varies across tasks, though its overall impact remains limited. 
For tasks like MMLU, which require multi-domain knowledge and exhibit high noise, filtering out lower-variance components improves performance by reducing irrelevant or conflicting information. In contrast, for tasks like GSM8K, which focus on mathematical reasoning with minimal domain conflict, retaining more components helps preserve crucial task-related knowledge, enhancing performance.
Despite these task-specific variations, the effect of adjusting the principal component ratio is constrained. In most cases, our method consistently outperforms traditional selection approaches, highlighting its robustness and effectiveness across different tasks.

\subsection{The Role of Graph Structure and Gradient-based Representations}

To assess the contributions of the two core components of our method—the structured graph-based framework and gradient-based knowledge representation—we conduct a comparative ablation experiment.
Specifically, we introduce two ablation settings: (1) \textbf{w/o graph},where knowledge is extracted solely via PCA and selected based on gradient similarity, and (2) \textbf{w/o gradient},where the gradient-based similarity measure is replaced with a BERT-based semantic similarity approach.

The results on the COT and FLAN datasets, shown in Table \ref{table:ablation}, demonstrate the necessity of both structured graph modeling and gradient-based representations.
While both components improve performance, the structured graph is especially critical.
As the number of selected knowledge elements increases, the graph-based approach consistently outperforms by capturing interdependencies between instructions, reducing redundancy, and optimizing knowledge transfer.
These findings highlight the effectiveness of our graph-based selection framework in enhancing instruction tuning.

\section{Related Work}
Recent studies~\cite{park2023trak, jainimproving} have explored using model gradients for instruction-tuning data selection, showing that gradients capture the informational content of training samples.
\citet{xialess} proposed a gradient similarity-based approach that selects data aligned with the validation set, achieving performance comparable to full-data instruction tuning using only 5\% of the data.
\citet{joaquin2024in2core} improved data selection for large models by leveraging the similarity between the training and validation sets in smaller models, enhancing selection efficiency.

However, these approaches focus on similarity and fail to capture the deeper interdependencies within the training data. Previous research~\cite{zhao2024beyond} found that instructions are not independent, but instead exhibit interdependencies. By leveraging these dependencies, instruction tuning performance can be improved. Moreover,~\citet{hubotter2024efficiently} highlighted that similarity-based selection overlooked these dependencies, limiting the effectiveness of instruction tuning.

Several studies~\cite{lu2023instag, bhattexperimental} enhance dataset complexity and diversity.
While modifying data distributions for generalization, they prioritize adaptability over domain-specific optimization and omit explicit modeling of relationships between training samples.

With the rise of domain-specific LLMs~\cite{ wu2023bloomberggpt, zhang2023xuanyuan}, instruction selection methods should account for dependencies in training data.
Unlike similarity-based approaches, graph-based models capture interdependencies, reduce redundancy, and enhance knowledge transfer.
This highlights the need for more effective selection strategies that extend beyond similarity, optimizing instruction tuning for specialized domains.
Graph-based models, like G2IS, offer a promising solution by capturing complex relationships, leading to improved performance.

\section{Conclusion}

We introduced G2IS, a gradient-based instruction selection method aimed at enhancing domain-specific instruction tuning by constructing structured gradient-based graphs. Unlike traditional methods, which rely on similarity measures, G2IS captures the joint distribution and interdependencies between instructions, leading to more efficient data selection. Our experiments on state-of-the-art models demonstrate that G2IS significantly outperforms conventional instruction tuning approaches, achieving superior results with only 1\% of the training data, especially on complex reasoning tasks like GSM8K and BBH. Notably, G2IS excels in selecting highly relevant data, showing that smaller, carefully curated subsets outperform larger, less relevant datasets. Ablation studies further confirm the critical role of both the structured graph framework and gradient-based representations in optimizing knowledge transfer and improving performance. These findings emphasize G2IS's potential to enhance task-specific learning, reduce data requirements, and enable more efficient instruction-tuning strategies, particularly in data-limited specialized domains.

\section{Limitations}

In this study, we adopt the LoRA method and compute gradients only for the LoRA layers, rather than the full model parameters, to reduce computational cost.While our method demonstrates strong performance across various instruction-tuning tasks, we have not yet evaluated whether computing full-parameter gradients could further enhance data selection effectiveness.
Additionally, due to computational constraints, our experiments are primarily conducted on 7B and 8B-scale models (e.g., Llama3-8B, Gemma-7B, Mistral-7B), and we have not yet tested our method on larger-scale LLMs (e.g., 13B, 65B, 175B).
In future work, we plan to extend our study to full-gradient computation and larger-scale models to more comprehensively assess the applicability and optimization potential of the G2IS method.
\bibliography{custom}

\appendix 
\twocolumn
\section{Appendix: Data Construction}
\label{appendix:data}

\subsection{Training Data Construction}

\begin{itemize}
    \item \textbf{Infinity-Instruct} \footnote{\url{https://huggingface.co/datasets/BAAI/Infinity-Instruct}}  
    We randomly selected 1 million instruction-tuning samples from the InfInstruct-3M core subset for training.
    
    \item \textbf{COT}~\cite{wei2022chain} \textbf{and FLAN-v2}~\cite{longpre2023flan}  
    We used the cleaned dataset provided by~\cite{xialess} to ensure data quality.
\end{itemize}

\subsection{Validation and Test Data Construction \& Evaluation Methods}

\begin{itemize}
    \item \textbf{BBH Dataset}~\cite{suzgun2022challenging}  
    \begin{itemize}
        \item Validation set: We used the original dataset’s development set, which consists of 81 samples.
        \item Evaluation method: 0-shot evaluation.
    \end{itemize}

    \item \textbf{GPQA Dataset}~\cite{rein2023GPQA}  
    \begin{itemize}
        \item Validation set: Since the original dataset does not include an official development set, we randomly selected 55 samples as the validation set, with the remaining samples used for testing.
        \item Evaluation method: 0-shot evaluation.
    \end{itemize}

    \item \textbf{GSM8K Dataset}~\cite{cobbe2021GSM8K}  
    \begin{itemize}
        \item Validation set: We randomly selected 100 samples from the training set as the validation set.
        \item Evaluation method: 5-shot evaluation.
    \end{itemize}

    \item \textbf{Math Dataset}~\cite{hendrycks2021measuring}  
    \begin{itemize}
        \item Validation set: We randomly selected 200 samples from the training set as the validation set.
        \item Evaluation method: 4-shot evaluation.
    \end{itemize}

    \item \textbf{MMLU Dataset}~\cite{hendryckstest2021}  
    \begin{itemize}
        \item Validation set: We used the original dataset’s dev set, which consists of 1,531 samples.
        \item Evaluation method: Default evaluation approach as specified in the original dataset.
    \end{itemize}
\end{itemize}

To ensure consistency and fairness in the evaluation process, we used the harness evaluation framework~\cite{eval-harness}, keeping all hyperparameters at their default settings.

\section{Appendix: Train Setup}
\label{appendix:train}
\subsection{Experimental Setup}

All experiments were conducted on an \textbf{A100-SMX4} GPU cluster.To ensure fair comparisons and reproducibility, we used the \textbf{Llama-Factory}~\cite{zheng2024llamafactory} for model training and followed the experimental configurations outlined in~\cite{xialess}. 

The model was fine-tuned using the \textbf{LoRA} method with a rank of 128, $\alpha=512$, and a dropout rate of 0.1.We employed \texttt{bf16} precision and used the \texttt{cosine} learning rate scheduler with an initial learning rate of $2\times10^{-5}$.The training process spanned 3 epochs, with a batch size of 2 per device, gradient accumulation steps set to 16, and a warm-up ratio of 0.3.The maximum sequence length was set to 2048 tokens.For optimization, we used the DeepSpeed ZeRO-2~\cite{rajbhandari2020zero} configuration.

\subsection{Warmup Strategy and Random Projection}

To effectively initialize the momentum of the Adam optimizer, we employed a \textbf{warmup} strategy.Specifically, we randomly selected 5000 samples from the training dataset and conducted 4 epochs of training to initialize the LoRA layer parameters and optimizer momentum.Given the high computational cost of tracking momentum at every training step, we approximate the model’s momentum when encountering new data by using the momentum accumulated after these 4 epochs.Since the model sizes are similar, we adopted hyperparameters similar to those used in~\cite{xialess}. 

Due to the significant magnification of gradients in the LoRA layer, we reduce the dimensionality of the LoRA gradients by applying Random Projection~\cite{ park2023trak}, as suggested by~\cite{xialess}, to project the gradients down to 8192 dimensions.

\subsection{Computational Cost Analysis}  

The primary computational cost of this experiment comes from gradient computation.However, since we adopt the LoRA  method and compute gradients \textbf{only for the LoRA layers}, the overall computational overhead is \textbf{significantly reduced} compared to full-parameter instruction tuning.Furthermore, the additional cost compared to existing gradient-based similarity methods is negligible.In the construction of the gradient-based graph, we employ \textbf{efficient sparse computation techniques}, and the entire process can be completed \textbf{on the CPU}, eliminating the need for additional GPU resources.Overall, \textbf{G2IS introduces minimal additional computational cost} while significantly improving the efficiency and accuracy of data selection, making it both practical and scalable.

\onecolumn
\section{Appendix:Gradient Walk Algorithm for Instruction Selection}
\label{alg:gradient-walk}
\begin{algorithm}[H]
\caption{Gradient Walk Algorithm for Instruction Selection}
\KwIn{$\mathcal{Z}$: Training dataset, $\mathcal{V}$: Validation dataset, $\delta$: Threshold for validation consistency, $k$: Principal component selection threshold (e.g., cumulative variance ratio), $\text{ratio}$: Percentage of training data to be selected}
\KwOut{$\mathcal{S}$: Selected instruction subset}

\textbf{Step 1: Compute Gradients} \\
\For{each validation sample $v \in \mathcal{V}$}{
    Compute validation gradient: $\nabla_{\text{val}}(v) = \nabla \mathcal{L}(v, \theta)$
}
\For{each training sample $z \in \mathcal{Z}$}{
    Compute training gradient: $\nabla_{\text{train}}(z) = \nabla \Gamma(z, \theta)$ \tcp{Momentum-adjusted gradient}
}

\textbf{Step 2: Extract Core Knowledge from Validation Set} \\
Perform PCA on $\{\nabla_{\text{val}}(v) | v \in \mathcal{V} \}$ \\
Select top-$k\%$ principal components based on cumulative variance ratio \\
Represent validation core knowledge as $K_{\mathcal{V}}$ \\
Normalize weights: $\alpha_i = \frac{\omega_i}{\sum_{j=1}^{k} \omega_j}$ so that $\sum_{i=1}^{k} \alpha_i = 1$

\textbf{Step 3: Gradient Walk Selection} \\
\For{each principal component $i$}{
    Compute selection budget: $N_{\text{select}, i} = \text{ratio} \times |\mathcal{Z}| \times \alpha_i$
}

Initialize $\mathcal{S} \leftarrow \emptyset$ \\

\For{each $v \in K_{\mathcal{V}}$}{
    Initialize $\mathcal{S'} \leftarrow \emptyset$ \\
    Choose $s \in \mathcal{Z}$ maximizing $\text{cos}(\nabla_{\text{train}}(s), v)$ \\
    $\mathcal{S'} \leftarrow \mathcal{S'} \cup \{s\}$

    \While{$|\mathcal{S'}| < N_{\text{select}, i}$}{
        Sort $z^*$ from $\mathcal{Z} \setminus \mathcal{S'}$ in descending order by $\text{cos}(\nabla_{\text{train}}(z^*), \nabla_{\text{train}}(s))$ \\
        Set \textit{found} = \textbf{False}

        \For{each sorted $z^*$}{
            \If{$\forall s' \in \mathcal{S'}, \text{cos}(\nabla_{\text{train}}(z^*), \nabla_{\text{train}}(s')) \geq 0$ \textbf{and} $\big| \text{cos}(\nabla_{\text{train}}(\mathcal{S'} \cup \{z^*\}), K_{\mathcal{V}}) \big| \geq \delta \cdot \big| \text{cos}(\nabla_{\text{train}}(\mathcal{S'}), K_{\mathcal{V}}) \big|$}{
                $\mathcal{S'} \leftarrow \mathcal{S'} \cup \{z^*\}$ \\
                Update $s = z^*$ \\
                Set \textit{found} = \textbf{True} \\
                \textbf{break for-loop}
            }
        }
        
        \If{\textbf{not} \textit{found}}{
            Choose $z^* = \arg\max_{z \in \mathcal{Z} \setminus \mathcal{S'}} \text{cos}(\nabla_{\text{train}}(z), K_{\mathcal{V}})$ \\
            Update $s = z^*$ \\
            $\mathcal{S'} \leftarrow \mathcal{S'} \cup \{z^*\}$ \\
        }
    }
    $\mathcal{S} \leftarrow \mathcal{S} \cup \mathcal{S'}$
}

\KwRet{$\mathcal{S}$}
\end{algorithm}


\end{document}